\pdfoutput=1 
\documentclass{JINST}

\title{COMPUTER AIDED DETECTION OF ORAL LESIONS ON
CT IMAGES}

\author{Shaikat Galib,$^a$ F. Islam,$^a$ M. Abir,$^{b}$ and
H. K. Lee$^a$\thanks{Corresponding author.}~\\
\llap{$^a$}Department of Mining and Nuclear Engineering\\ 
Missouri University of Science and Technology\\ 
301 W 14th St, Rolla, MO-65409, USA\\
\llap{$^b$}Idaho National Laboratory\\ 
 P.O. Box- 1625, MS-6188,  Idaho Falls, ID 83415-6188, USA\\
E-mail: \email{leehk@mst.edu}}

\abstract{Oral lesions are important findings on computed tomography (CT) images. In this study, a fully automatic method to detect oral lesions in mandibular region from dental CT images is proposed. Two methods were developed to recognize two types of lesions namely (1) Close border (CB) lesions and (2) Open border (OB) lesions, which cover most of the lesion types that can be found on CT images. For the detection of CB lesions, fifteen features were extracted from each initial lesion candidates and multi layer perceptron (MLP) neural network was used to classify suspicious regions. Moreover, OB lesions were detected using a rule based image processing method, where no feature extraction or classification algorithm were used. The results were validated using a CT dataset of 52 patients, where 22 patients had abnormalities and 30 patients were normal. Using non-training dataset, CB detection algorithm yielded 71\% sensitivity with 0.31 false positives per patient. Furthermore, OB detection algorithm achieved 100\% sensitivity with 0.13 false positives per patient. Results suggest that, the proposed framework, which consists of two methods, has the potential to be used in clinical context, and assist radiologists for better diagnosis. }

\keywords{Oral lesions; Computed Tomography (CT); Computer Aided Diagnosis; Computer Aided Detection; Algorithms; Medical image analysis; Artificial Neural Network}

\begin{document}

\section{Introduction}\label{sec:intro}

Computer Aided Detection (CAD) is a technology that interprets and explains digital images. In the field of medicine, CAD systems are usually developed for extracting useful information from medical images (e.g. Radiography, MRI, Tomography, Ultrasound, PET, SPECT etc.) and providing a second opinion to doctors. Study shows that a well-designed CAD system can increase medical doctors' performance and, therefore, reduce incorrect actions and diagnosis time \cite{bib1}. Several CAD systems have been implemented commercially for detecting breast cancer, lung cancer, colon cancer, Alzheimers disease, among others.

In this paper, we are proposing a CAD system that is expected to detect lesions from oral CT images in mandibular region. According to the report of National Cancer Institute, oral cancer accounts for 2.5\% of all cancers in the United States \cite{bib2}. Research shows, if oral cancer is detected in early stages, the death rate can be reduced to 10\% - 20\% while later stages lead to 40\% - 65\% mortality. 

Oral CT scan is commonly used for treatment planning of orthodontic issues, temporomandibular joint disorder diagnosis, correct placement of dental implants, evaluating the jaw, sinuses, nerve canals and nasal cavity; detecting, assessing and treating jaw tumors and many more. Interpretation of dental CT images are challenging because image modalities are often poor due to noise, contrast is low and artifacts are present; topology is complicated; teeth orientation is arbitrary and lacks clear lines of separation between normal and abnormal regions [4]. Moreover, the inspection of the CT scan requires dedicated training and dentists' time. Furthermore, the diagnosis may vary from dentist to dentist and experience plays a vital role in correct judgment and conclusion of diagnosis. Besides, some early lesions may not be visible clearly to the human eye. These issues and their probable solutions for a better diagnostic environment are the primary motivation of this work.

Oral CT images contain both maxilla and mandible of oral anatomy. Lesions in the mandibular region only were studied in this work, and a method to detect its abnormalities was proposed. The framework consists of two algorithms, one is for detection of Close border (CB) lesion (Type I) and the other is for detection of Open border (OB) lesion (Type II). Image processing involved in these two algorithms were implemented in 2D CT slice images and final decision and lesion marking was done by analyzing the results in a 3D CT volume. This paper reports the methodology and evaluates the results, to validate the goal of a CAD system.

\section{Related works}\label{sec:RW}

Computerized dental treatment systems and clinical decision support systems have seen success in recent years. Many diverse CAD systems were developed for diagnosing different oral diseases. Based on most researched cases, we have broadly classified the research areas into three different categories: caries diagnosis, bone density diagnosis and lesion/bone defect diagnosis. Following sections describe the categories in short. 
 
\subsection{Caries diagnosis} 
Computer aided caries diagnosis technology is perhaps the most studied category in the dental radiograph analysis field. The Logicon System (Carestream Dental LLC, Atlanta, GA) is a well-known technology for caries detection \cite{bib3}. The output of the software is in graphical form, which shows whether the area in question is a sound tooth, or is decalcified or carious. Tracy et al. (2011) \cite{bib1} studied the performance the of the CAD system where twelve blinded dentists reviewed 17 radiographs. The group concluded that, by using the CAD support, diagnosis of caries increased from 30\% to 69\%. 
Likewise, Firestone et al. (1998) \cite{bib5} investigated the effect of a knowledge-based decision support system (CariesFinder, CF) on the diagnostic performance and therapeutic decisions. The study involved 102 approximal surface radiographic images and 16 general practitioners to identify the presence of caries and whether restoration was required. Their study showed that when the dental practitioners were equipped with a decision support system, their ability to diagnose dental caries correctly increased significantly. 
Similarly, Olsen et al. (2009) \cite{bib6} proposed a computerized diagnosis system that aimed to give feedback about the presence and extent of caries on the surface of teeth. Their method gave both qualitative and quantitative opinion to dental practitioners, by using digital images and a graphical user interface. 

\subsection{Bone density diagnosis}
Kavitha et al. (2012) \cite{bib7} proposed a CAD system that measures the cortical width of the mandible continuously to identify women with low bone mineral density (BMD) from dental panoramic images. The algorithm was developed using support vector machine classifier where images of 60 women were used for system training and 40 were used in testing. Results showed that the system is promising for identifying low skeletal BMD. 
Muramatsu et al. (2013) \cite{bib8} also proposed a similar work for measuring mandibular cortical width with a 2.8 mm threshold. The algorithm showed 90\% sensitivity and 90\% specificity. 
Reddy et al. (2011) \cite{bib9} developed a CAD method to differentiate various metastatic lesions present in the human jawbones from Dental CT images. They developed a method to find most discriminative texture features from a region of interest, and compared support vector machine (SVM) and neural network classifier for classification among different bone groups. They have achieved an overall classification accuracy of 95\%, and concluded that artificial neural networks and SVM are useful for classification of bone tumors. 

\subsection{Lesion/bone defect diagnosis} 
Shuo Li et al. (2007) \cite{bib10} developed a semi-automatic lesion detection framework for periapical dental X-rays using level set method. The algorithm was designed to locate two types of lesions: periapical lesion (PL) and bifurcation lesion (BL). Support vector machine was used for segmentation purpose. The algorithm automatically locates PL and BL with a severity level marked on it. Stelt et al. (1991) \cite{bib11} also developed a similar algorithm to detect periodontal bone defects. They have used image processing techniques to find the lesions that were artificially introduced into the radiographs. Their system was designed to decrease interobserver variability and time-dependent variability. 

The work in this research was focused on lesion/bone defect diagnosis on CT images. The proposed method is a fully automatic scheme to highlight suspicious regions regardless of the type of a disease. In the following sections, materials used for this work and methodology of the algorithm are discussed, and finally the results are evaluated.

\section{ Materials}

Total oral CT scans of 52 patients were investigated, where 22 patients had abnormalities and 30 patients had normal appearances. The CT scans were obtained using a 495-slice spiral CT scanner (Vatech, PaX - i3D) with tube voltage of 90 KV, current of 10 mA and slice thickness of 0.2mm. The 2D slice data was reconstructed with a 512 x 512 matrix. A dentist verified normal and abnormal aspects of the images. These images were provided by Vatech Co., Ltd, South Korea, and were supplied in 16-bit RAW format.

\section{ Methodology }

\begin{figure}[tbp]
\centering
\includegraphics[width=0.5\textwidth]{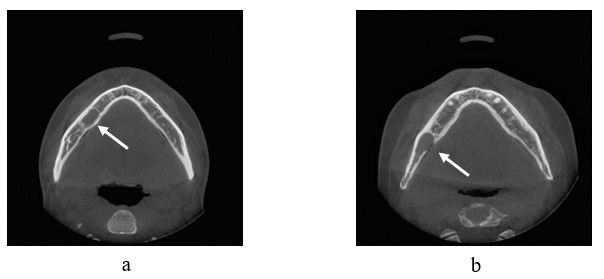} 
\caption{Illustration of two types of abnormality (a) Close border lesion, (b) Open border lesion
}\label{fig:2type4}  
\end{figure}

\begin{figure}[tbp]
\centering
\includegraphics[width=0.8\textwidth]{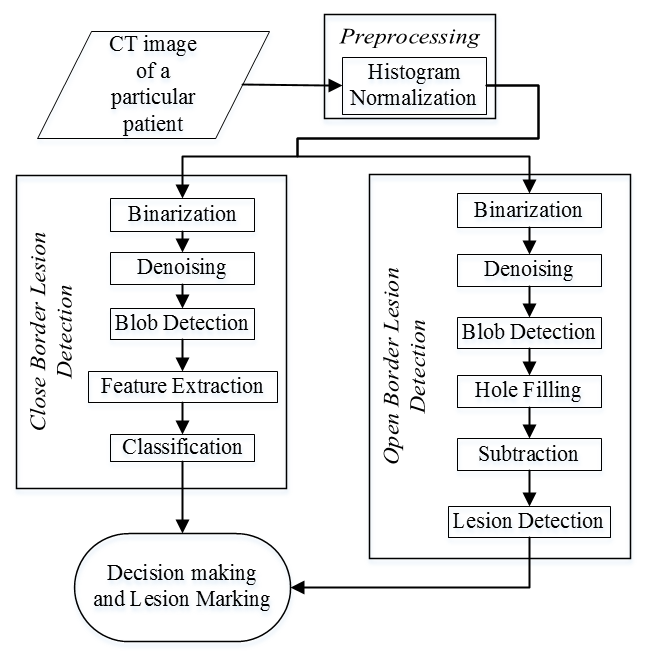} 
\caption{Architecture of the proposed method
}\label{fig:Algorithm} 
\end{figure}

This section elaborates the method of the CAD system (Figure \ref{fig:Algorithm}). The framework consists of two algorithms for detection of two types of lesions: Close border lesions (Type I) and Open border lesions (Type II) (Figure \ref{fig:2type4}). This classification is not done on a medical basis but is used for research purpose only. In addition, the goal of this CAD scheme was not to detect a specific syndrome, rather any kind of abnormality regardless of the diseases type. Therefore, the classification of lesions considered here served as a generalized grouping of all type of abnormalities that could be found in dental CT images. A Close border lesion is defined as one that has a well-defined boundary around the lesion, and a Open border problem is one that has a broken boundary line around the lesion. Methods for detecting these two types of lesions are described in the later sections.

\subsection{Preprocessing}

To enhance the contrast of the image, histogram normalization was carried out according to equation \ref{eq:preprocessing}:

\begin{equation}\label{eq:preprocessing}
I_{out(x,y)}=  255 \bigg( \frac{I_{in(x,y)}-I_{in,min}}   {I_{in,max}-I_{in,min}}\bigg)
\end{equation}

\subsection{Detection of Close border (CB) Lesion (Type I) }

A CB lesion in dental CT images can be understood as a low-intensity region surrounded by a relatively high-intensity boundary. These regions can be circular, elliptical or any irregular shape. The problematic regions were usually neither very large nor very small in size, and had a unique texture and statistical properties. These regions were visually homogenous and could be identified in several slices for a particular patient. CB lesion detection algorithm was designed in three classic stages: initial lesion candidate detection, feature extraction and classification. These steps are described in sequential sections.

\subsubsection{Initial lesion candidate detection }

The purpose of initial lesion candidate detection is to detect all the possible areas that could be a potential lesion. Potential CB lesions can be considered as blobs in a digital CT image. Blob can be realized as regions in images that differ in properties compared to its surrounding area. The sensitivity of this blob detection process was high so the chances of missing out a possible lesion candidate were low in this stage.

\paragraph{Binarization.} \label{Bin}
The binary image was computed from original gray scale image using Maximum Entropy threshold method \cite{bib12}, which is an automatic threshold process. Calculating appropriate threshold value for binarization is vital for the performance of detection because inaccurate threshold could eliminate possible lesion candidates. This particular method was chosen due to its binarization accuracy and implementation simplicity. 

\paragraph{Denoising.}
To eliminate noises from the binary image, morphological closing operation  \cite{bib13} was carried out. Structuring element for the operation was chosen in such a way that it would fill holes of size up to 5mm diameter. Holes larger than 5mm were considered as a possible lesion.

\paragraph{Blob detection.}

\begin{figure}[tbp]
\centering
\includegraphics[width=.8\textwidth]{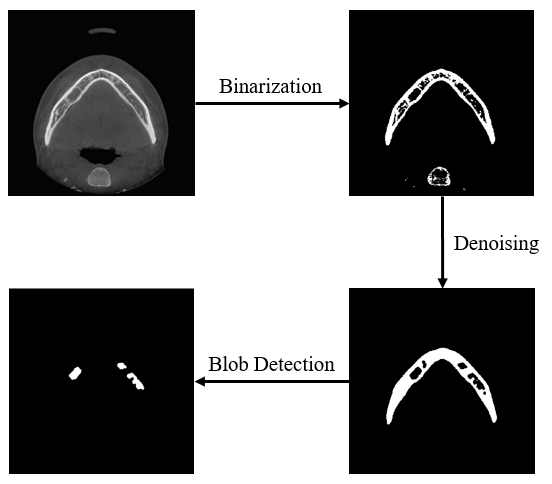}   
\caption{Illustration of Initial Lesion Candidate Detection process
}\label{fig:ILCD}
\end{figure}

Finally, all possible lesion candidates were filtered out from denoised image by applying blob detection \cite{bib14} method, which was developed by Haralick et al. (1992). All blobs larger than 5mm in size were detected. In addition, the anatomy present in the bottom 1/3 of the image area was less likely to represent a lesion, therefore, detected blobs in this area were rejected. Figure \ref{fig:ILCD} illustrates the initial lesion candidate detection process.

\subsubsection{Feature extraction}
\begin{figure}[tbp]
\centering
\includegraphics[width=.8\textwidth]{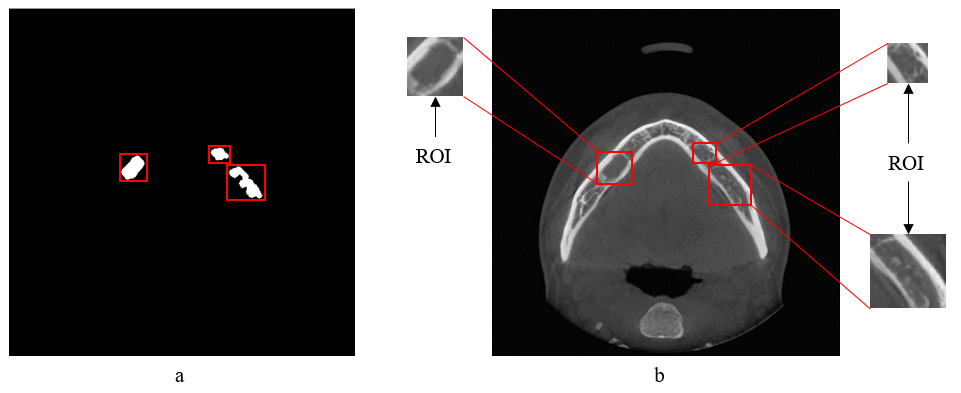} 
\caption{Illustration of ROI extraction process. (a) Bounding rectangles calculated around each blob, (b) Bounding rectangles were superimposed over the intensity image to crop out the ROIs.}\label{fig:ROI_extraction}  
\end{figure}

The reason of feature extraction is to acquire sufficient information about an image. Once all the initial lesion candidates were detected, a bounding rectangle around each blob were calculated (Figure \ref{fig:ROI_extraction}(a)). The rectangles were then superimposed over the intensity image in the same position relative to the detected blobs (Figure \ref{fig:ROI_extraction}(b)). Then image sections inside each rectangle were cropped out for feature extraction and considered as Region of Interest (ROI). 

Three types of features were extracted from each ROI. These features are commonly known as First order statistics, Second order statistics and Image moments. 

First Order Statistics are the standard statistical measures of gray level values of a ROI. Mean, standard deviation, skewness and kurtosis were calculated from each ROI.

Second order statistics, computed from Grey-Level Co-occurrence Matrices (GLCM) \cite{bib15}, are a good measure of image texture. This feature set was developed by Haralick et al. (1973) and is used in many real world applications for its reasonable performance in discriminating different image textures. Four second order statistics features were calculated namely: contrast, homogeneity, energy and entropy. For the computation of GLCM, only horizontal position of neighboring pixels were considered. 

Image moments provide a special kind of weighted average of image pixels intensities. This measure usually provide attractive information such as geometry information of objects in an image. Hu (1962) \cite{bib16} developed seven image moments which are independent of position, size, orientation and parallel projection. These seven moments were calculated for extracting geometry information from ROIs. 

Therefore, a total of fifteen (First order statistics: 4, Second order statistics: 4, moments: 7) feature vectors were calculated from each ROI.

\subsubsection{Classification}

\begin{figure}[tbp]
\centering
\includegraphics[width=.9\textwidth]{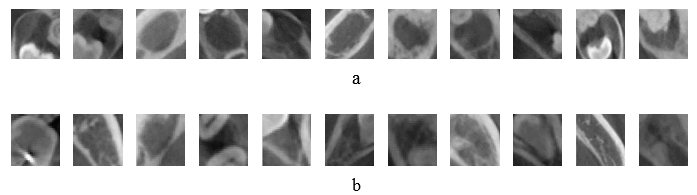}  
\caption{Illustrations of training samples (a) Samples of lesion ROI, (b) Samples of normal ROI.}\label{fig:training} 
\end{figure}

Artificial Neural Network (ANN) was used for classification of data. Specifically, Multilayer Perceptron (MLP) neural network was employed for classification between two groups: Lesion and Normal. The network was trained with backpropagation learning \cite{bib17}. Sample ROIs were used to train the network, where both normal and abnormal examples are present (Figure \ref{fig:training}). Data was collected manually by using blob detection and feature extraction algorithms. The neural network had a configuration of fifteen input layer nodes for fifteen feature vectors, ten hidden layer nodes, and two output layer nodes. The trained network was then used for unseen data classification.

Finally, a lesion area was marked on the intensity image if the ROI was classified as a lesion class.

\subsection{Detection of Open border (OB) lesion (Type II)}

\begin{figure}[tbp]
\centering
\includegraphics[width=1.0\textwidth]{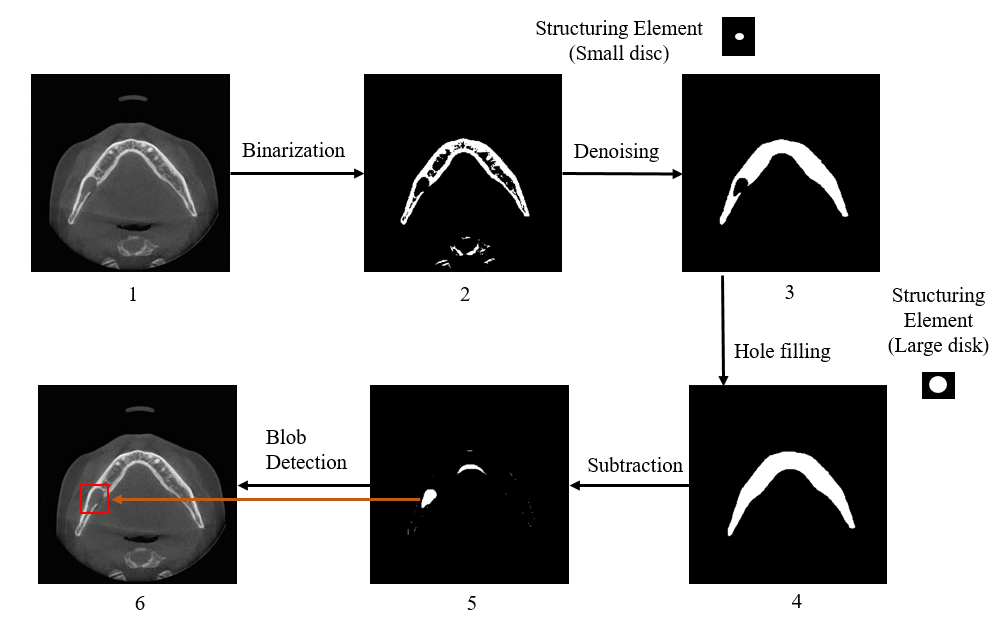}   
\caption{Illustration of Open border lesion detection process. (1) Grayscale CT slice, (2) Binary image, (3) Noise reduction, (4) Large hole filling, (5) Subtraction result of image 3 from image 4., (6) Lesion area marking on the intensity image. 
}\label{fig:BDmethod}
\end{figure}

Open border problems cannot be solved by blob detection method, as lesions had no distinctive closed border separating it from the background. However, this problem was solved by using morphological image processing operations. The method is schematically shown in Figure \ref{fig:BDmethod}: 

\begin{enumerate}
\item [Step 1]

A grayscale CT slice. (Figure \ref{fig:BDmethod}(1)). 

\item [Step 2]

Binary image of the grayscale CT slice (Figure \ref{fig:BDmethod}(2)). See section \ref{Bin} for binarization.
\item [Step 3]

Morphological closing operation was carried out on the binary image to remove noise and artifacts from it. Structuring element for this closing operation was designed in such a way that it can remove holes or openings smaller than 5mm diameter. Moreover, any anatomy present in the bottom 1/3 of the image was eliminated from further consideration. (Figure \ref{fig:BDmethod}(3)). 

\item [Step 4]

Morphological closing operation was carried out once more to fill large sized holes or openings (i.e. closing with larger disk shaped structuring element). The design of the structuring element was such that it can fill holes up to 30mm of diameter (Figure \ref{fig:BDmethod}(4)). 

\item [Step 5]

The output image from step 3 was then subtracted from the image of step 4. The resulting image contained any architectural defect present in the intensity image. In this stage, blobs were detected that were larger than 5mm diameter and smaller than 30mm diameter. If any blob found within the specified range, then it was considered as a probable lesion (Figure \ref{fig:BDmethod}(5)). 

\item [Step 6]

Finally, the detected blob centroid was shown on the intensity image (Figure \ref{fig:BDmethod}(6)).
\end{enumerate}

\section{Results and Discussions}

In this section, results of the two algorithms are discussed separately. In the first section, results of CB lesion detection algorithm have been evaluated, and results of OB lesion detection algorithm will be evaluated in the later section. 

In our CT dataset, 22 patients contained oral lesions where 10 patients had CB lesions only, 8 patients had  OB lesions only and 4 patients had both types of lesions. We had also considered 30 patients who did not have any lesions (normal cases). Thus, 52 CT cases composed the dataset to evaluate the lesion detection framework.

Free Response Receiver Operating Characteristics (FROC) curves were used for evaluating the CAD system. Sensitivity and false positive per patient were the parameters used for FROC curves where,

\begin{equation}
Sensitivity=  \frac{TPs}{TPs+FNs}
\end{equation}

\begin{equation}
False\; positive\; per\; patient = \frac{FPs}{FPs+TNs}
\end{equation}

Where TP: True Positive, FP: False Positive, TN: True Negative and FN: False Negative. 

\subsection{Evaluation of Close border lesion detection algorithm}

\begin{figure}[tbp]
\centering
\includegraphics[width=0.7\textwidth]{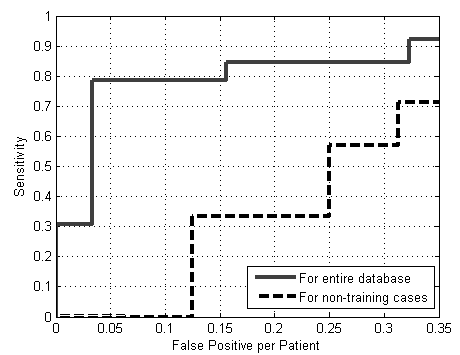}   
\caption{Comparison of FROC curves in Close border lesion detection method. The solid curve indicates the results of  the scheme when entire Close border dataset was used for both training and testing (14 abnormal and 30 normal patients). The dotted curve indicates the results when the testing set was not used for training.
}\label{fig:CBL} 
\end{figure}

To validate the proposed CB lesion detection method, we divided the CB patient cases into two sets- one for training and the other for testing. The training set contained 7 abnormal and 15 normal patients, and the testing set also contained 7 abnormal and 15 normal patients. Results of the algorithm are presented in Figure \ref{fig:CBL}. The testing set yielded a maximum of 71\% sensitivity with 0.31 false positives per patient. The FROC curve shows results at different operating points where the algorithm can be functioned. However, when all the 14 CB cases and all 30 normal cases were used for training, and the same dataset was used for testing, maximum 92\% sensitivity was achieved with 0.32 false positives per patient.

The algorithm performance was observed to be lower for those lesions that do not have a well-defined boundary around it. It is still a challenging task to detect lesions with unclear appearances. Moreover,  lesion size also had an impact to the performance of the algorithm. Our method intends to achieve high detection accuracy on oral lesions sized between 7.5mm to 20mm diameter. Detection accuracy decreased for lesions larger than 25mm or smaller than 7mm diameter. 

For training, only 7 abnormal and 15 normal example cases were used, which was a very small quantity of data to produce reliable results. However, the algorithm still performed in an acceptable manner by indicating 71\% sensitivity. Upon availability of additional example cases, the algorithm is expected to predict results with better accuracy and could be used in clinical context.

\subsection{Evaluation of Open border lesion detection algorithm}

\begin{figure}[tbp]
\centering
\includegraphics[width=0.7\textwidth]{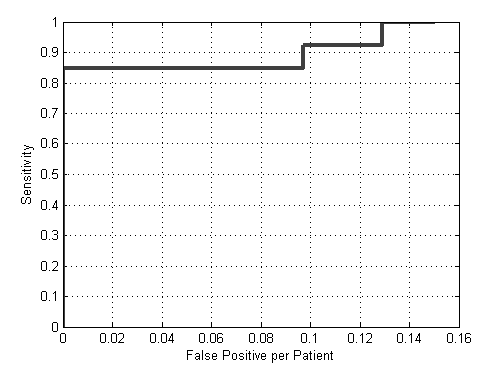}  
\caption{FROC Curve of Open border detection method
}\label{fig:bd}
\end{figure}

The OB algorithm yielded better results than the CB algorithm. It achieved 85\% sensitivity without any false positives on the OB dataset of 12 abnormal and 30 normal patients. Moreover, the algorithm was able to achieve 100\% sensitivity with just 0.13 false positives per patient (See figure \ref{fig:bd})

The system was designed to detect lesions of diameter between 10mm - 25mm. For the detection of larger or smaller sized lesions, more example data and research is needed.

In this method, no feature extraction or classification technique was used. Instead, a rule based process was implemented. The method was validated by dentists' opinion about healthy or lesion regions. Figure \ref{fig:bd} shows the performance of the method at different operating points.

Both of the systems were implemented on C++ language. It took 7 - 10 seconds for executing the algorithms together for a single patient. The computer used for the testing had core i7, 4.00 GHz processor and 16 GB of RAM.

Oral CAD algorithms are relatively new and emerging technology in the field of medical image analysis. Most of the work done in the past were for serving a specific purpose, as discussed in the related works section. The design of our CAD scheme is a more generalized procedure of lesion detection framework and it provides a different justification as well. Moreover, the image database used in this study is entirely different than others. Therefore, for comparison, the authors could not find any similar CAD scheme either in the literature or in the industry and our method can be considered as a novel approach of solving a new set of problems.

\section{Conclusion}
\label{concl}

An approach to detect oral lesions in mandible region on CT images have been proposed in this work. Two types of lesions, Close border lesions and Open border lesions, were identified. They covered most of the problems that could be found on oral CT images. Results suggest that this method is capable of detecting CB lesions with 71\% sensitivity at 0.31 false positives per patient and OB lesions with 100\% sensitivity at 0.13 false positives per patient.

	Some of the limitations of the method include that: it cannot detect lesions brighter than its surroundings, lack of a well-defined boundary around a lesion decreases the detection accuracy and too large or too small lesions achieved less detection sensitivity. However, availability of more example images are expected to improve the results and overcome the limitations.
    
	The work done in this research focused on the mandibular region of the oral anatomy. Therefore, future research possibilities in this CAD framework includes: detection of lesions in maxilla, detection of brighter lesions and improve the detection accuracy for lesions with unclear appearances.

\acknowledgments

The authors are thankful to  Vatech (Value Added Technology CO., LTD), South Korea, for providing the CT images and sponsoring the project.



\begin{thebibliography}{9}

\bibitem{bib1}
K.D. Tracy, B.A. Dykstra, D.C. Gakenheimer, J.P. Scheetz, S. Lacina, W.C. Scarfe and A.G. Farman,
\emph{Utility and effectiveness of computer-aided diagnosis of dental caries},
\href{http://www.scopus.com/inward/record.url?eid=2-s2.0-80053417378&partnerID=40&md5=7e7fcc92f866d22076acb333db552992}
{\emph{General Dentistry} \textbf{59} (2011) 136}


\bibitem{bib2}
National Cancer Institute,
\emph{Cancer statistics},
\href{http://seer.cancer.gov/statfacts/html/oralcav.html}
{\emph{Last accessed: April, 2015} }.

\bibitem{bib3}
D.C. Gakenheimer,
\emph{The efficacy of a computerized caries detector in intraoral digital radiography},
\href{http://www.scopus.com/inward/record.url?eid=2-s2.0-0036634481&partnerID=40&md5=a49328f0cc0977045a38e50a8201bade}
{\emph{Journal of the American Dental Association} \textbf{133} (2002) 883}.





\bibitem{bib5}
A.R. Firestone, D. Sema, T.J. Heaven, R.A. Weems,
\emph{The Effect of a Knowledge-Based, Image Analysis and Clinical Decision Support System on Observer Performance in the Diagnosis of Approximal Caries from Radiographic Images},
\href{http://www.scopus.com/inward/record.url?eid=2-s2.0-0031606021&partnerID=40&md5=a684c7803c4ac033b13d885dc41b450a}
{\emph{Caries Research} \textbf{32} (1998) 127}.



\bibitem{bib6}
G.F. Olsen, S.S. Brilliant, D. Primeaux, K. Najarian,
\emph{An image-processing enabled dental caries detection system},
\href{http://www.scopus.com/inward/record.url?eid=2-s2.0-67650723782&partnerID=40&md5=4d6ebf33a9e1e0624ef4e002a69ea743}
{\emph{2009 ICME International Conference on Complex Medical Engineering, CME 2009} \textbf {2009}} 



\bibitem{bib7}
M.S. Kavitha, A. Asano, A. Taguchi, T. Kurita, M. Sanada,
\emph{Diagnosis of osteoporosis from dental panoramic radiographs using the support vector machine method in a computer-aided system},
\href{http://www.scopus.com/inward/record.url?eid=2-s2.0-84855758521&partnerID=40&md5=5b61d9d2c609a44d1ee794cd71b396c6}
{\emph{BMC Medical Imaging} \textbf{12} (2012)}


\bibitem{bib8}
C. Muramatsu, T. Matsumoto, T. Hayashi, T. Hara, A. Katsumata, X. Zhou, Y. Iida, M. Matsuoka, T. Wakisaka, H. Fujita,
\emph{Automated measurement of mandibular cortical width on dental panoramic radiographs},
\href{http://www.scopus.com/inward/record.url?eid=2-s2.0-84888287489&partnerID=40&md5=72ed038e957ded2c9cc1cc96aa09cbed}
{\emph{International Journal of Computer Assisted Radiology and Surgery} \textbf{8} (2013) 877}


\bibitem{bib9}
T.K. Reddy, N. Kumaravel,
\emph{Multi resolution based texture analysis of jaw bone lesions},
\href{http://www.scopus.com/inward/record.url?eid=2-s2.0-79953810336&partnerID=40&md5=b3351bb06fee735fd68d5acd77ab435b}
{\emph{European Journal of Scientific Research} \textbf{51} (2011) 415}

\bibitem{bib10}
S. Li, T. Fevens, A. Krzyzak, C. Jin, S. Li,
\emph{Semi-automatic computer aided lesion detection in dental X-rays using variational level set},
\href{http://www.scopus.com/inward/record.url?eid=2-s2.0-34249031807&partnerID=40&md5=9aa5ea5ace390ecc4cec78387e76fcd0}
{\emph{Pattern Recognition} \textbf{40} (2007) 2861}


\bibitem{bib11}
P.F. van der Stelt, W.G.M. Geraets,
\emph{Computer-aided interpretation and quantification of angular periodontal bone defects on dental radiographs},
\href{http://www.scopus.com/inward/record.url?eid=2-s2.0-0026142732&partnerID=40&md5=3e8d9e2183993ad9b50da0d05d1443fa}
{\emph{IEEE Transactions on Biomedical Engineering} \textbf{38} (1991) 334}


\bibitem{bib12}
Jagat Narain Kapur, Prasanna K Sahoo and Andrew KC Wong,
\emph{A new method for gray-level picture thresholding using the entropy of the histogram},
\href{http://www.sciencedirect.com/science/article/pii/0734189X85901252}
{\emph{Computer vision, graphics, and image processing} \textbf{29} (1985) 273}



\bibitem{bib13}
Rafael C. Gonzalez and Richard E. Woods, 
\emph{Digital Image Processing}, 
Addison-Wesley Longman Publishing Co., Inc., 1992.



\bibitem{bib14}
Robert M. Haralick and Linda G. Shaprio, 
\emph{Computer and Robot Vision}, 
Addison-Wesley Longman Publishing Co., Inc., 1992.


\bibitem{bib15}
Hu Ming-Kuei,
\emph{Visual pattern recognition by moment invariants},
\href{http://dx.doi.org/10.1109/TIT.1962.1057692}
{\emph{IRE Transactions on Information Theory} \textbf{8} (1962) 179}

\bibitem{bib16}
Y. LeCun, L. Bottou, G. Orr, K.R. Müller,
\emph{Efficient BackProp},
\href{http://dx.doi.org/10.1007/3-540-49430-8_2}
{\emph{Springer Berlin Heidelberg} \textbf{1524} (1998) 9}


\bibitem{bib17}
R.M. Haralick, K. Shanmugam, and Its'Hak Dinstein,
\emph{Textural Features for Image Classification},
\href{http://dx.doi.org/10.1109/TSMC.1973.4309314}
{\emph{Systems, Man and Cybernetics, IEEE Transactions on} \textbf{SMC-3} (1973) 610}


\end{thebibliography}
\end{document}